\documentclass[11pt]{article}

\usepackage[final]{acl}

\usepackage{times}
\usepackage{latexsym}
\usepackage[T1]{fontenc}
\usepackage[utf8]{inputenc}
\usepackage{microtype}
\usepackage{inconsolata}
\usepackage{graphicx}
\usepackage{booktabs}
\usepackage{amsmath}
\usepackage{amssymb}
\usepackage{multirow}
\usepackage{url}

\title{Quality-Aware Self-Correcting Speech Translation on an Edge Device}

\author{Zubair Ajmal Farooq \\
  School of CS and EE \\
  University of Surrey \\
  United Kingdom \\
  \texttt{zubair5ajmal@gmail.com} \\\And
  Diptesh Kanojia \\
  Institute for People-Centred AI \\
  University of Surrey \\
  United Kingdom \\
  \texttt{d.kanojia@surrey.ac.uk} \\}

\begin{document}
\hypersetup{
  pdftitle={Quality-Aware Self-Correcting Speech Translation on an Edge Device},
  pdfauthor={Zubair Ajmal Farooq, Diptesh Kanojia}
}
\maketitle

\begin{abstract}
We present a fully offline speech-to-speech translation pipeline that runs on a Jetson Nano (4\,GB) and corrects its own weak translations without retraining. A Whisper-tiny ASR feeds an Opus-MT translator; multilingual BERT cosine similarity acts as a Quality Estimation (QE) gate, triggering a second-pass correction when confidence falls below a pre-defined threshold $\tau$. We compare three correction methods: QE reranking (M1), Minimum Bayes-Risk decoding (M2), and constrained beam search (M3). On 1{,}012 FLORES-200 sentences (English-Spanish), M2 at $\tau{=}0.90$ produces statistically significant improvements over greedy decoding on BLEU ($+0.67$, $p{<}0.001$), ChrF ($+0.51$, $p{<}0.001$), and COMET ($+0.0020$ at $N{=}3$, $p{=}0.002$); M1 yields no significant gains, and M3 is significantly worse than baseline ($p{>}0.99$). Our central finding is that QE functions effectively as a \emph{gate} but poorly as a \emph{ranker}: removing the QE model from candidate selection (M1$\to$M2) does not hurt quality and frees 680\,MB from the critical path. Using a gain-to-edit ratio adapted from the post-editing-effort literature, we further show that smaller candidate pools ($N{=}3$) yield more surgical corrections with better semantic adequacy, while larger pools ($N{=}10$) maximise lexical reward. We release the system and demonstrate live translation across six language pairs\footnote{\url{https://github.com/surrey-nlp/speech-translation_edge_device}}.
\end{abstract}

\section{Introduction}

Speech translation increasingly relies on cloud services that expose audio, transcripts, and intent. For healthcare, legal, and low-bandwidth deployments this is unacceptable. Edge devices offer a privacy-preserving alternative, but cascaded speech translation pipelines (ASR\,$\to$\,MT\,$\to$\,TTS) compound errors at each stage, and small models compatible with edge hardware are particularly susceptible. We ask: can a sub-\$150 edge device translate speech, detect its own weak translations, and correct them, fully offline, without retraining?

We answer affirmatively. Our system runs end-to-end on a Jetson Nano (4\,GB) and uses a lightweight Quality Estimation (QE) model to trigger a second-pass correction over an N-best candidate pool. We evaluate three correction strategies under a $4\times 3$ grid of QE thresholds and beam sizes, with paired bootstrap significance testing on 1{,}012 FLORES-200 sentences.

Our contributions are:
\begin{itemize}\itemsep0pt
\item A complete offline speech translation system with QE-triggered self-correction running on 4\,GB of RAM (\S\ref{sec:system}).
\item Statistically significant quality improvements via Minimum Bayes-Risk reranking (M2) on BLEU, ChrF, and COMET, with paired bootstrap evidence (\S\ref{sec:results}).
\item The finding that QE serves as a reliable trigger but an unreliable candidate ranker, with practical implications for memory footprint on edge hardware (\S\ref{sec:gate-not-ranker}).
\item An adapted gain-to-edit ratio (\S\ref{sec:ger}) that connects per-sentence COMET improvements to edit distance, providing an analytical lens on the lexical-vs-semantic trade-off observed across beam pool sizes.
\item A latency analysis (\S\ref{sec:latency}) demonstrating that correction overhead on this hardware is dominated by model reload, not inference, and that a single-load architecture roughly halves per-utterance latency (from \textasciitilde 82\,s to \textasciitilde 41\,s).
\end{itemize}

\section{Related Work}
\label{sec:related}

\paragraph{Edge MT and on-device translation.}
End-to-end on-device speech translation has typically required server-class hardware or compressed cascades \citep{whisper2022,opusmt2020}. Edge deployments of single-stage MT components are common (mobile translators, USB sticks), but composed pipelines with on-device quality control are rare. Our work targets the smallest publicly available NVIDIA edge SKU with 4\,GB unified memory.

\paragraph{Quality Estimation for MT.}
Reference-free QE \citep{specia2018qe} estimates translation quality without parallel references. We use multilingual BERT cosine similarity as a lightweight QE signal; stronger neural QE models such as COMET-Kiwi \citep{cometkiwi2022} offer better correlation with human judgement but were too large for our memory budget. We adopt the position that even a noisy QE signal can be useful as a gate, even where it fails as a ranker.

\paragraph{Self-correction and reranking.}
Minimum Bayes-Risk decoding \citep{kumar2004mbr} selects the candidate with highest peer-consensus utility from an N-best list, and has been shown to outperform maximum-likelihood decoding when paired with strong utility functions \citep{freitag2022mbr}. QE-based reranking selects the candidate with the highest reference-free quality score \citep{fernandes2022quality}. Constrained beam search forces decoding to include specified terms \citep{post2018constrained}, and has been used in domain adaptation and terminology enforcement. We apply all three to the self-correction setting and quantify their relative performance under bootstrap testing.

\paragraph{Edit-distance-based effort metrics.}
Edit-distance measures of MT quality and effort date to the early 1990s \citep{frederking1994,king1996}. The Translation Edit Rate (TER) \citep{snover2006ter} and its human-targeted variant HTER \citep{snover2006hter,specia2010pe} are the most widely used. Post-editing effort metrics including post-edit distance and time-to-edit \citep{daems2017post} normalise quality or effort over edit counts to characterise human correction work. Our gain-to-edit ratio (\S\ref{sec:ger}) adapts this lineage to the \emph{automatic} self-correction setting: rather than measuring human effort against a reference, we measure neural quality gain (COMET) per unit of edit distance between two MT versions of the same source.

\section{System}
\label{sec:system}

The pipeline is a cascade of four components, each chosen for its memory footprint under a 4\,GB unified-memory ceiling.

\paragraph{ASR.} OpenAI Whisper-tiny (\textasciitilde 39\,M parameters) transcribes English audio. We use the public PyTorch implementation with greedy decoding.

\paragraph{MT.} Helsinki-NLP Opus-MT \texttt{en-es} (\textasciitilde 78\,M parameters) translates the transcript. First-pass decoding uses greedy search ($N{=}1$); second-pass correction uses beam search with $N\in\{3, 5, 10\}$.

\paragraph{QE.} Multilingual BERT base \citep{devlin2019bert} produces sentence embeddings for source and target; we use cosine similarity as the QE score $s\in[-1,1]$. mBERT was selected over COMET-Kiwi for memory reasons (\textasciitilde 680\,MB vs.\ \textasciitilde 2\,GB peak). We do not claim cosine similarity is the best QE signal; we claim it is adequate as a binary trigger.

\paragraph{TTS.} eSpeak-ng synthesises Spanish output. Output latency is dominated by audio synthesis rather than text processing.

\paragraph{Memory architecture.}
Under sequential loading (one model resident at a time), each stage incurs a load/unload cost dwarfing inference. Under a single-load architecture all four models reside concurrently in \textasciitilde 3.6\,GB, well within the 4\,GB ceiling; per-utterance latency drops accordingly (\S\ref{sec:latency}).

\paragraph{Multilingual extension.}
Although our evaluation is English-Spanish, the system extends to six language pairs (en\,$\leftrightarrow$\,es, fr, de) by swapping the Opus-MT model and the TTS voice; live demonstration confirms operation across all pairs at comparable steady-state latency. We restrict quantitative evaluation to en-es for tractability and depth.

\section{Self-Correction Methods}
\label{sec:methods}

When the QE score $s$ falls below threshold $\tau$, the system generates $N$ alternative translations and selects one via one of three methods.

\paragraph{M1: QE reranking.} The $N$ candidates are scored against the source using the same mBERT cosine measure used as the gate. The highest-scoring candidate is selected.

\paragraph{M2: Minimum Bayes-Risk (MBR).} Each candidate is scored by its mean ChrF similarity to the other $N{-}1$ candidates (peer consensus). The candidate with highest mean similarity is selected. MBR does not consult the source or the QE model after candidate generation.

\paragraph{M3: Constrained beam search (CBS).} Content words longer than three characters in the greedy first-pass output are extracted as soft anchors. Beam search is repeated under a constrained decoding objective that boosts hypotheses containing these anchors; the resulting candidates are then reranked by mBERT QE.

\section{Experimental Setup}
\label{sec:setup}

\paragraph{Data.} We use the English--Spanish \texttt{devtest} split of FLORES-200: 1{,}012 sentences drawn from English Wikimedia sources (Wikinews, Wikijunior and Wikivoyage), balanced across topical domains (language codes \texttt{eng\_Latn}\,$\to$\,\texttt{spa\_Latn}). For the ASR component, we use FLEURS \citep{conneau2023fleurs} English audio with paired transcripts.

\paragraph{Grid.} We sweep $\tau\in\{0.75, 0.80, 0.85, 0.90\}$ and $N\in\{3, 5, 10\}$ across all three correction methods, producing 36 configurations. The greedy baseline ($N{=}1$, no correction) is computed once and used as the comparison point for every configuration.

\paragraph{Metrics.} We report sacreBLEU \citep{post2018sacrebleu}, ChrF \citep{popovic2015chrf}, and COMET-22 \citep{rei2022comet}. Significance is assessed by paired bootstrap resampling \citep{koehn2004bootstrap} with $n{=}1000$ resamples; we report one-tailed $p$-values for the hypothesis that the correction method improves over the baseline.

\paragraph{Hardware.} All ASR/MT/QE runs were performed on an NVIDIA Jetson Nano 4\,GB Developer Kit (Maxwell GPU, 128 CUDA cores, 4\,GB LPDDR4) with a 40\,mm PWM cooling fan. COMET evaluation was performed on Google Colab CPU (the model exceeds available memory on the Nano). Identical model weights and hypothesis files were used in both environments.

\section{Results}
\label{sec:results}

\subsection{Aggregate Quality}

Table~\ref{tab:headline} reports the strongest configuration for each correction method at $\tau{=}0.90$ together with bootstrap significance. M2 produces significant improvements on BLEU and ChrF, and on COMET at $N{=}3$ (\S\ref{sec:tradeoff}); M1 produces no significant change; M3 is significantly worse than baseline. The full $4\times3$ grid across all $(\tau, N)$ combinations appears in Appendix~\ref{app:fullgrid}.

\begin{table}[t]
\centering
\small
\begin{tabular}{lrrr}
\toprule
Method & $\Delta$BLEU & $\Delta$ChrF & $\Delta$COMET \\
\midrule
M1 (QE-rerank)  & $-0.14$ & $-0.11$ & $-0.0004$ \\
M2 (MBR)        & $\mathbf{+0.67}^{***}$ & $\mathbf{+0.51}^{***}$ & $\mathbf{+0.0011}$ \\
M3 (CBS)        & $-0.96^{\dagger}$ & $-0.86^{\dagger}$ & $-0.0061^{\dagger}$ \\
\bottomrule
\end{tabular}
\caption{Aggregate deltas over greedy baseline at $\tau{=}0.90$, $N{=}10$. Paired bootstrap, $n{=}1000$. $^{***}$: $p{<}0.001$. $^{\dagger}$: significantly \emph{worse} than baseline ($p{>}0.99$, one-tailed). $\Delta$COMET for M2 at $N{=}10$ is not individually significant ($p{=}0.107$); see \S\ref{sec:tradeoff}.}
\label{tab:headline}
\end{table}

\subsection{Beam-Size Trade-off: Lexical vs Semantic}
\label{sec:tradeoff}

A direct corollary of running the full grid is that different beam pool sizes win different metrics (Table~\ref{tab:tradeoff}). At $\tau{=}0.90$:
\begin{itemize}\itemsep0pt
\item \textbf{BLEU and ChrF} peak at $N{=}10$ ($+0.67$ and $+0.51$, both $p{<}0.001$).
\item \textbf{COMET} peaks at $N{=}3$ ($+0.0020$, $p{=}0.002$); $N{=}5$ and $N{=}10$ produce smaller, non-significant gains.
\end{itemize}
The full grid (Appendix~\ref{app:fullgrid}) shows that this pattern holds robustly: M2 outperforms M1 and M3 at every combination where correction actually fires ($\tau \geq 0.85$).

\begin{table}[t]
\centering
\small
\begin{tabular}{lrrr}
\toprule
$N$ & $\Delta$BLEU & $\Delta$ChrF & $\Delta$COMET \\
\midrule
3  & $+0.59^{***}$ & $+0.37^{***}$ & $\mathbf{+0.0020^{**}}$ \\
5  & $+0.47^{**}$  & $+0.32^{**}$  & $+0.0012$ \\
10 & $\mathbf{+0.67^{***}}$ & $\mathbf{+0.51^{***}}$ & $+0.0011$ \\
\bottomrule
\end{tabular}
\caption{M2 (MBR) corrections at $\tau{=}0.90$ across beam pool sizes. Paired bootstrap, $n{=}1000$. $^{**}$: $p{<}0.01$. $^{***}$: $p{<}0.001$. $\Delta$COMET at $N{=}5$ and $N{=}10$ are not significant at $\alpha{=}0.05$ (one-tailed $p{=}0.082$ and $0.107$).}
\label{tab:tradeoff}
\end{table}

This pattern is consistent with the structure of MBR: the consensus candidate is chosen for maximal peer agreement on lexical surface, not semantic adequacy against the source. With $N{=}10$, the consensus has more room to drift toward a safe, lexically common phrasing that rewards $n$-gram overlap against the reference but may dilute semantic precision. With $N{=}3$, the candidate pool is small and the consensus stays closer to the original greedy output, preserving semantic adequacy at the cost of lexical novelty.

\subsection{Gain-to-Edit Ratio}
\label{sec:ger}

To make the lexical-versus-semantic trade-off observable per sentence rather than only at corpus level, we adapt the edit-distance lineage from post-editing-effort metrics \citep{snover2006hter,specia2010pe,daems2017post} to the automatic self-correction setting. We define the gain-to-edit ratio (GER) for each sentence whose output the correction changed (so that TER${>}0$) as
\begin{equation}
\mathrm{GER} = \frac{\Delta \mathrm{COMET}}{\mathrm{TER}(\text{baseline}, \text{corrected})}
\end{equation}
where $\Delta\mathrm{COMET}$ is the per-sentence COMET improvement of the corrected output over the baseline and TER is the Translation Edit Rate \citep{snover2006ter} between the two MT versions. GER is not a new metric; it is an adapted ratio whose components and form derive from prior work. We use it as an analytical lens. A higher GER indicates surgical correction (small edit, real semantic gain); a lower GER indicates blunt rewrites that perturb the output without improving adequacy.

Table~\ref{tab:ger} reports median GER for M2 across $N$. The ranking aligns with the COMET column of Table~\ref{tab:tradeoff}: $N{=}3$ has the highest median GER, $N{=}5$ the lowest, $N{=}10$ intermediate. The mean GER for $N{=}10$ is negative ($-0.0019$) while its median is positive ($+0.0048$); this asymmetry indicates a long tail of large, low-gain edits at larger $N$. Median is the more robust summary.

\begin{table}[t]
\centering
\small
\begin{tabular}{lrr}
\toprule
$N$ & GER (median) & Changed \\
\midrule
3   & $\mathbf{+0.0053}$ & 485 \\
5   & $+0.0023$ & 542 \\
10  & $+0.0048$ & 580 \\
\bottomrule
\end{tabular}
\caption{Median per-sentence GER for M2 at $\tau{=}0.90$. Changed = number of sentences (out of 1{,}012) where the correction altered the baseline output; GER is computed over these. This is a subset of the sentences where QE triggered correction (Table~\ref{tab:fullgrid}).}
\label{tab:ger}
\end{table}

GER thus surfaces a property that aggregate BLEU/ChrF/COMET cannot. The lexical and semantic metrics disagree about which $N$ is best; GER explains why both are right: at smaller $N$, MBR makes targeted edits that preserve semantic adequacy; at larger $N$, MBR is more willing to substantially rewrite the output, which can win on lexical surface but introduces noise.

\begin{figure}[t]
\centering
\includegraphics[width=\columnwidth]{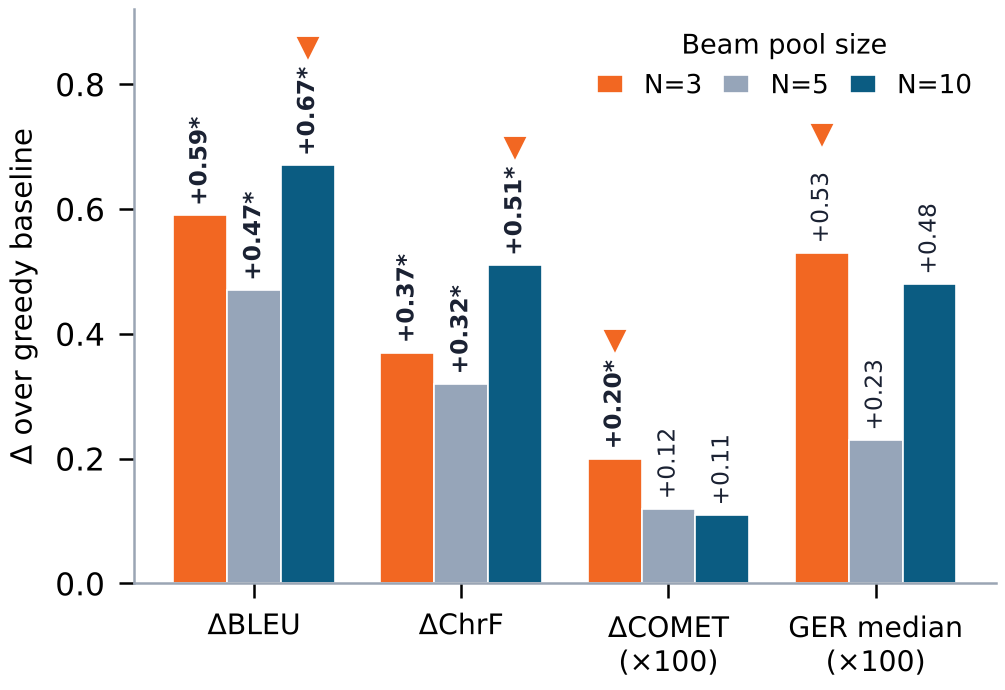}
\caption{Improvement over greedy baseline across four metrics for M2 (MBR) at $\tau{=}0.90$. $^{*}$: $p{<}0.05$ (paired bootstrap, $n{=}1000$); $\blacktriangledown$: best $N$ per metric. $\Delta$COMET and GER median are scaled $\times 100$ for joint visibility. Lexical metrics (BLEU, ChrF) peak at $N{=}10$; semantic metric (COMET) and per-sentence efficiency (GER median) peak at $N{=}3$.}
\label{fig:pareto}
\end{figure}

Figure~\ref{fig:pareto} integrates the four metrics (BLEU, ChrF, COMET, GER median) for M2 at $\tau{=}0.90$ across the three beam pool sizes, making the per-metric winners explicit.

\subsection{QE as Gate, Not Ranker}
\label{sec:gate-not-ranker}

A second observation from Table~\ref{tab:headline} is that removing the QE model from the candidate-selection step while retaining it as the trigger (the difference between M1 and M2) does not degrade quality. M2 in fact outperforms M1 on every metric. This is consistent with prior findings that reference-free QE signals at the segment level are noisier discriminators among similar candidates than between clearly good and clearly bad translations \citep{fernandes2022quality}. The practical implication for edge deployment is direct: once the gate has fired, the QE model can be unloaded for the duration of MBR selection, freeing \textasciitilde 680\,MB of resident memory.

This finding generalises beyond our specific system. Wherever a cascaded MT pipeline uses a lightweight QE model both for triggering and for ranking, our results suggest splitting these roles: keep the QE model as a trigger gate, but let the candidates themselves vote on the selection.

\subsection{M3 Failure}

At $\tau{=}0.90$, CBS underperforms baseline on every metric and at every $N$, with $p{>}0.99$ in the one-tailed bootstrap test. We attribute this to a structural problem: the anchors are extracted from the greedy first-pass output, the very output that the QE gate has just flagged as low-confidence. Constraining the second pass to preserve those anchors forces the system to retain whatever errors the trigger fired on. CBS would likely be useful in a different role (e.g., terminology enforcement, named-entity preservation) but seems inappropriate as a quality-driven correction strategy for noisy first-pass output.

\section{Latency Analysis}
\label{sec:latency}

Per-utterance latency on the Jetson Nano under sequential loading averages approximately 82\,s, with model reload accounting for the majority of stage cost. Correction adds approximately 23.9\,s, dominated by the load/unload cycle around MT Pass 2 rather than by the beam search itself.

Under a single-load architecture, all four models remain resident for the duration of the session. We measured this configuration on the Nano: peak resident memory was 3.6\,GB (within the 4\,GB ceiling), and per-utterance latency for one $\sim$10\,s en-es utterance was approximately 41\,s (including first-utterance load amortisation). The MBR self-correction overhead reduces from 23.9\,s to 13.2\,s once model reload is removed from the critical path.

\begin{figure}[h]
\centering
\includegraphics[width=\columnwidth]{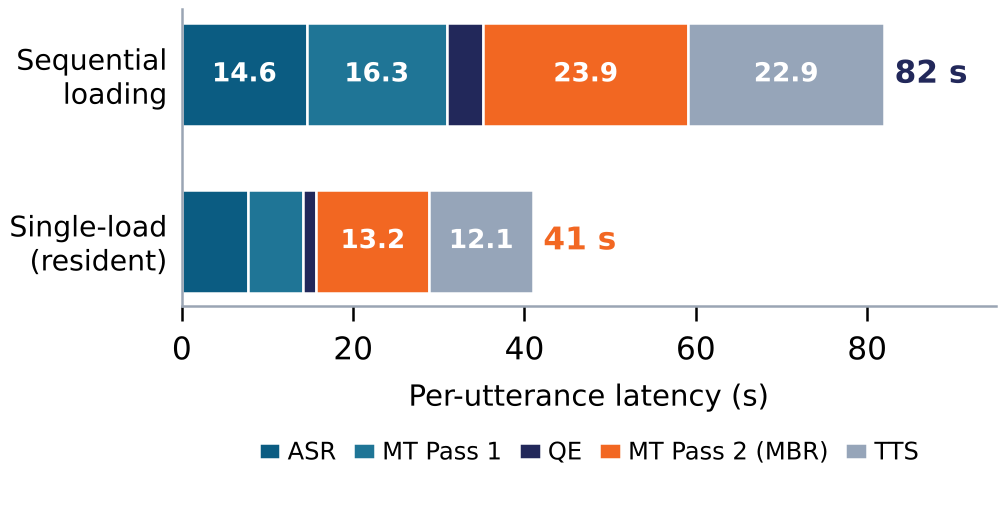}
\caption{Per-stage latency on the Jetson Nano for one $\sim$10\,s en-es utterance. Sequential-load (top, $\sim$82\,s) is dominated by per-stage model reload; single-load resident (bottom, $\sim$41\,s) eliminates reload at the cost of higher peak memory.}
\label{fig:latency}
\end{figure}

Critically, the bottleneck under the original architecture is memory, not the algorithm. The Nano's 4\,GB ceiling \emph{forced} the sequential architecture: under those constraints, the user-perceived latency reflects memory bandwidth rather than the computational cost of self-correction. Hardware with a modestly larger memory budget (e.g., Jetson Orin Nano 8\,GB, Raspberry Pi 5 with swap, mobile-class SoCs) would run the same pipeline at near-real-time latency without any change to the algorithms reported here.

\section{Discussion}

Our findings can be summarised as follows. (1) QE-triggered MBR self-correction produces statistically significant improvements on lexical metrics and on COMET at $N{=}3$. (2) The QE model is valuable as a trigger but not as a ranker, which has direct memory-footprint consequences for edge deployments. (3) The lexical-versus-semantic trade-off across beam pool sizes is explainable in terms of MBR's optimisation target and is observable at sentence level through GER. (4) CBS is structurally unsuited to correction on flagged output. (5) Edge latency is RAM-bound, not compute-bound.

\section{Conclusion}
We demonstrate that an edge device can translate speech, detect its own weak translations, and correct them, fully offline, without retraining. The required machinery (MBR self-correction over a small candidate pool, QE used as a gate rather than as a ranker, models held resident under a modest memory budget) is straightforward to implement and produces measurable quality improvements. The principal obstacle to deploying conversational on-device speech translation today is not the cost of self-correction algorithms but the cost of RAM.

\section*{Limitations}
First, evaluation is restricted to English-Spanish; we demonstrate but do not evaluate the system across the additional five language pairs supported by the same architecture. Second, the QE signal is multilingual BERT cosine similarity, which is a weak proxy for semantic adequacy. Stronger reference-free QE models (COMET-Kiwi) would likely produce a more discriminative gate but exceed available memory; quantising or distilling such models for edge use is an open direction. Third, FLORES-200 \texttt{devtest} is a written-text benchmark with controlled domains; performance under spontaneous speech with disfluencies, code-switching, or noisy acoustic conditions has not been measured. Fourth, COMET-22 is a learned metric whose own biases propagate into GER; using multiple independent quality estimators would strengthen claims about semantic adequacy.
\bibliography{custom}

\appendix

\section{Full Grid Search Results}
\label{app:fullgrid}

Table~\ref{tab:fullgrid} reports the complete $4 \times 3$ grid search over QE thresholds $\tau \in \{0.75, 0.80, 0.85, 0.90\}$ and beam sizes $N \in \{3, 5, 10\}$ for all three correction methods on 1{,}012 FLORES-200 sentences. The internal baseline values are BLEU $26.09$ and ChrF $54.85$; parenthesised values are deltas from these baselines. Trigger rates indicate the proportion of sentences for which QE fired correction at each threshold.

\begin{table*}[t]
\centering
\small
\begin{tabular}{ccrrrrrrr}
\toprule
$\tau$ & $N$ & Trig\% & BLEU-M1 & BLEU-M2 & BLEU-M3 & ChrF-M1 & ChrF-M2 & ChrF-M3 \\
\midrule
0.75 & 3  & 0.1\%  & 26.10 ($+0.01$) & 26.10 ($+0.01$) & 26.10 ($+0.01$) & 54.85 ($+0.00$) & 54.85 ($+0.00$) & 54.87 ($+0.02$) \\
0.75 & 5  & 0.1\%  & 26.10 ($+0.01$) & 26.10 ($+0.01$) & 26.10 ($+0.01$) & 54.87 ($+0.02$) & 54.85 ($+0.00$) & 54.87 ($+0.02$) \\
0.75 & 10 & 0.1\%  & 26.10 ($+0.01$) & 26.10 ($+0.01$) & 26.10 ($+0.01$) & 54.86 ($+0.01$) & 54.87 ($+0.02$) & 54.86 ($+0.01$) \\
\midrule
0.80 & 3  & 1.3\%  & 26.10 ($+0.01$) & 26.10 ($+0.01$) & 26.11 ($+0.02$) & 54.85 ($+0.00$) & 54.85 ($+0.00$) & 54.87 ($+0.02$) \\
0.80 & 5  & 1.3\%  & 26.10 ($+0.01$) & 26.09 ($+0.00$) & 26.11 ($+0.02$) & 54.86 ($+0.01$) & 54.84 ($-0.01$) & 54.87 ($+0.02$) \\
0.80 & 10 & 1.3\%  & 26.09 ($+0.00$) & 26.10 ($+0.01$) & 26.10 ($+0.01$) & 54.85 ($+0.00$) & 54.87 ($+0.02$) & 54.86 ($+0.01$) \\
\midrule
0.85 & 3  & 22.7\% & 26.23 ($+0.14$) & 26.34 ($+0.25$) & 26.22 ($+0.13$) & 54.93 ($+0.08$) & 54.99 ($+0.14$) & 54.95 ($+0.10$) \\
0.85 & 5  & 22.7\% & 26.14 ($+0.05$) & 26.26 ($+0.17$) & 26.22 ($+0.13$) & 54.84 ($-0.01$) & 54.94 ($+0.09$) & 54.85 ($+0.00$) \\
0.85 & 10 & 22.7\% & 26.11 ($+0.02$) & 26.40 ($+0.31$) & 26.02 ($-0.07$) & 54.84 ($-0.01$) & 55.04 ($+0.19$) & 54.78 ($-0.07$) \\
\midrule
0.90 & 3  & 86.2\% & 26.45 ($+0.36$) & 26.68 ($+0.59$) & 25.94 ($-0.15$) & 55.04 ($+0.19$) & 55.22 ($+0.37$) & 54.65 ($-0.20$) \\
0.90 & 5  & 86.2\% & 26.16 ($+0.07$) & 26.56 ($+0.47$) & 25.68 ($-0.41$) & 54.85 ($+0.00$) & 55.17 ($+0.32$) & 54.39 ($-0.46$) \\
0.90 & 10 & 86.2\% & 25.96 ($-0.13$) & \textbf{26.76 ($+0.67$)} & 25.14 ($-0.95$) & 54.74 ($-0.11$) & \textbf{55.35 ($+0.50$)} & 53.99 ($-0.86$) \\
\bottomrule
\end{tabular}
\caption{Full $4 \times 3$ grid search results over 1{,}012 FLORES-200 sentences. Internal baselines: BLEU $26.09$, ChrF $54.85$. M1 = QE-Rerank, M2 = MBR, M3 = CBS. Parenthesised values are deltas from baseline, computed from the rounded scores shown, so they may differ by $\pm 0.01$ from the unrounded deltas reported in the main text. Bold marks the strongest M2 configuration reported in the main body.}
\label{tab:fullgrid}
\end{table*}

\end{document}